\documentclass[sigconf]{acmart}

\copyrightyear{2025}
\acmYear{2025}
\setcopyright{cc}
\setcctype{by}
\acmConference[CIKM '25]{Proceedings of the 34th ACM International Conference on Information and Knowledge Management}{November 10--14, 2025}{Seoul, Republic of Korea}
\acmBooktitle{Proceedings of the 34th ACM International Conference on Information and Knowledge Management (CIKM '25), November 10--14, 2025, Seoul, Republic of Korea}\acmDOI{10.1145/3746252.3761084}
\acmISBN{979-8-4007-2040-6/2025/11}
\usepackage{graphicx} 

\AtBeginDocument{%
  }

\usepackage{hyperref}       % hyperlinks
\usepackage{url}            % simple URL typesetting
\usepackage{booktabs}       % professional-quality tables
\usepackage{amsfonts}       % blackboard math symbols
\usepackage{nicefrac}       % compact symbols for 1/2, etc.
\usepackage{microtype}      % microtypography
\usepackage{xcolor}         % colors
\usepackage{natbib}
\usepackage{graphicx}
\usepackage[table]{xcolor} % 
\usepackage{colortbl} 

\usepackage{amsthm,amsmath}
\usepackage{mathrsfs}
\usepackage{enumerate}
\usepackage{subfigure}
\usepackage{float}
\usepackage{bbding}
\usepackage{nicematrix}
\usepackage{arydshln}
\usepackage{balance}
\usepackage{multirow}

\usepackage{enumitem}

\newcommand{\stitle}[1]{\vspace{1mm}\noindent\textbf{#1}}

\begin{document}

% Modified based on KDD：
% 1. 15 datasets --> 13 datasets on MoleculeNet
% 2. Deleted D-MPNN (supervised) baseline
\title{Unified Molecule Pre-training with Flexible 2D and 3D Modalities: Single and Paired Modality Integration}

% not a genuine decoder
% a (end-to-end) seq2seq model
\author{Tengwei Song}
\authornote{This work was completed during a research visit to Singapore Management University.}
\affiliation{%
  \institution{Computational Bioscience Research Center, King Abdullah University of Science and Technology}
  \city{Jeddah}
  \country{Saudi Arabia}
}
\email{songtengwei@gmail.com}

\author{Min Wu}
\affiliation{%
  \institution{Institute for Infocomm Research, A*STAR}
  \country{Singapore}}
\email{wumin@i2r.a-star.edu.sg}

\author{Yuan Fang}
\authornote{Corresponding author.}
\affiliation{%
  \institution{School of Computing and Information Systems, Singapore Management University}
  \country{Singapore}
}
\email{yfang@smu.edu.sg}

\renewcommand{\shortauthors}{Tengwei Song, Min Wu, and Yuan Fang}

\begin{abstract}
Molecular representation learning plays a crucial role in advancing applications such as drug discovery and material design. Existing work leverages 2D and 3D modalities of molecular information for pre-training, aiming to capture comprehensive structural and geometric insights. However, these methods require paired 2D and 3D molecular data to train the model effectively and prevent it from collapsing into a single modality, posing limitations in scenarios where a certain modality is unavailable or computationally expensive to generate. To overcome this limitation, we propose FlexMol, a flexible molecule pre-training framework that learns unified molecular representations while supporting single-modality input. Specifically, inspired by the unified structure in vision-language models, our approach employs separate models for 2D and 3D molecular data, leverages parameter sharing to improve computational efficiency, and utilizes a decoder to generate features for the missing modality. This enables a multistage continuous learning process where both modalities contribute collaboratively during training, while ensuring robustness when only one modality is available during inference. Extensive experiments demonstrate that FlexMol achieves superior performance across a wide range of molecular property prediction tasks, and we also empirically demonstrate its effectiveness with incomplete data. Our code and data are available at \url{https://github.com/tewiSong/FlexMol}.
\end{abstract}




% \received{20 February 2025}
% \received[revised]{12 March 2025}
% \received[accepted]{5 June 2025}

\maketitle

\section{Introduction}
Molecular representation learning has become a cornerstone for applications in drug discovery \cite{drug_discovery_2024_1,drug_dis_2}, material science \cite{Chen2021}, and other scientific domains \cite{NEURIPS2023_8bd31288}. A central challenge in this field is how to effectively leverage both 2D molecular graphs and 3D geometric conformations. These two modalities offer complementary information: 2D graphs capture the chemical connectivity \cite{feng2024unicorn,luo2023transformerunderstand2d,yu2024multimodal}, while 3D geometries provide spatial and electronic details essential for understanding molecular interactions \cite{3dinfomax,du2023molecule,zhou2023unimol}.

% \begin{figure}
%     \centering
%     \includegraphics[width=\linewidth]{fig/intro.eps}
%     \caption{Two categories of models that integrate both 2D and 3D molecule modalities, and their respective advantages and drawbacks.}
%     \label{fig:intro}
% \end{figure}

%The first type of model separately learns 2D and 3D representations of molecular information using a 2D model and a 3D model, respectively. The 2D and 3D representations are then used for downstream tasks: the 2D representation is used for 2D tasks, and the 3D representation is used for 3D tasks. Alternatively, the 2D representation can be transformed into a 3D representation through an SE(3)-equivariant SDE transformation for 3D downstream tasks, or the 3D representation can be transformed into a 2D representation through an SE(3)-invariant SDE transformation for 2D downstream tasks. The second type of model uses a unified SE(3)-equivariant encoder to model both 2D and 3D molecular features, producing a fused 2D & 3D representation. This unified representation is then used for downstream tasks in either the 2D or 3D domain.

\begin{figure*}
    \centering
    \includegraphics[width=0.9\linewidth]{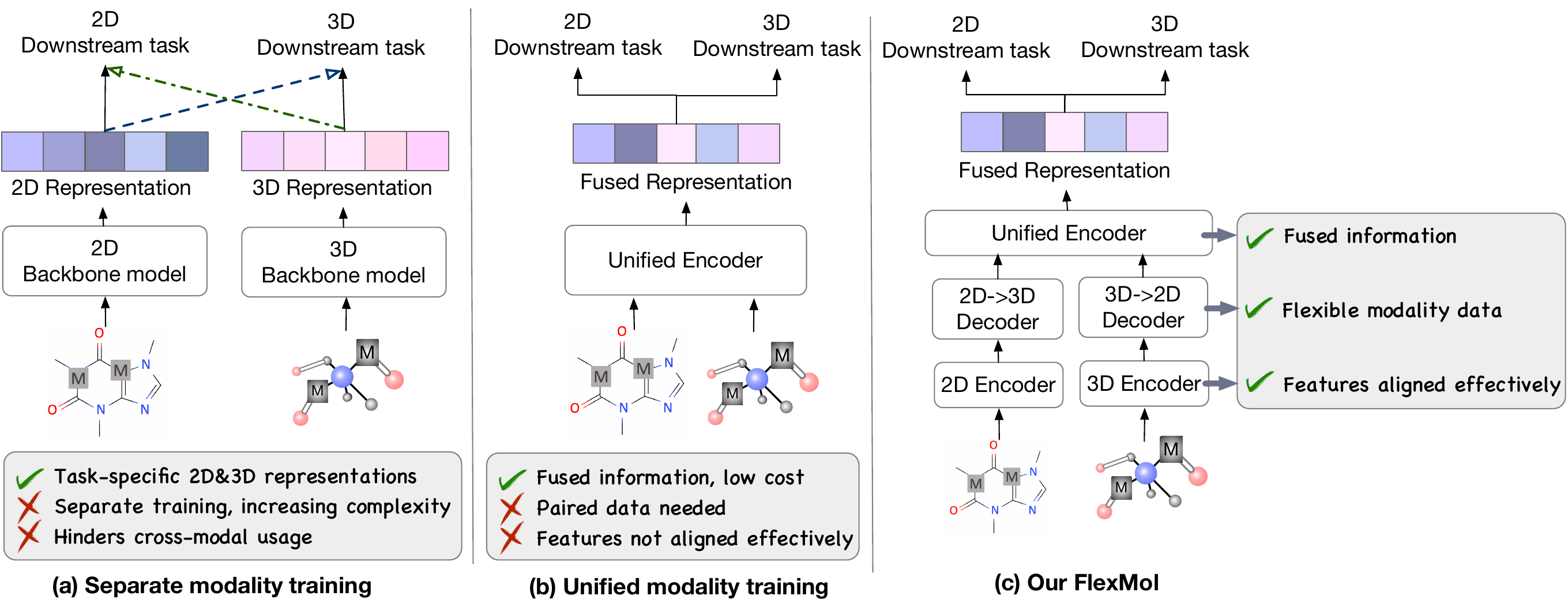}
    \vspace{-2mm}
    \caption{(a) \& (b) Two categories of models that integrate both 2D and 3D molecule modalities, and their respective advantages and drawbacks; (c) Our proposed FlexMol framework.}
    \label{fig:intro_ours}
\end{figure*}

\stitle{Limitations of Prior Work.} To learn from both 2D and 3D molecular data, current methods can be broadly divided into two categories, which are illustrated in Figure~\ref{fig:intro_ours}(a) and (b), respectively. 

The first category involves separate 2D and 3D outputs, such as GraphMVP \cite{graphmvp} and 
MoleculeSDE \cite{molsde}, where distinct models are trained independently on each modality. This approach allows each model to specialize, often improving accuracy for tasks relying on modality-specific features. 
When downstream tasks require cross-modality prediction, these models typically rely on SE(3)-equivariant Stochastic Differential Equation (SDE) to convert 2D representations into 3D or SE(3)-invariant SDE to transform 3D representations into 2D.
However, predicting 3D downstream tasks may also require 2D information, and vice versa. Since the 2D and 3D representations are modeled separately, the model cannot effectively leverage information across modalities.
%Additionally, combining outputs into a cohesive representation is complex, and paired 2D and 3D data is often difficult to obtain.

The second category aims to unify 2D and 3D molecular representations within a single model, offering computational efficiency and better integration of 2D and 3D features. However, without a proper alignment of the two modalities, the model may struggle to capture complementary information. Additionally, most of these approaches are not capable of handling unpaired data, such as UnifiedMol \cite{unified2d3d_kdd}, which relies on the joint use of paired 2D and 3D data (i.e., having both 2D graph and 3D conformation for each molecular) during pre-training.  Even when some models can handle single-modality data, it will reduce to a less effective single-modality model, such as Transformer-M \cite{luo2023transformerunderstand2d} and MolBlend \cite{yu2024multimodal}.

% \begin{figure}
%     \centering
%     \includegraphics[width=0.8\linewidth]{fig/intro_ours.eps}
%     \caption{Our FlexMol framwwork.}
%     \label{fig:intro_ours}
% \end{figure}

% \vspace{-1.5mm} 
\stitle{Our Work.} To address these limitations, we propose a novel framework called \emph{FlexMol}, which integrates the advantages of both approaches while mitigating their challenges, as illustrated in Figure~\ref{fig:intro_ours}(c). Specifically, first, to ensure that information from different modalities is effectively fused while maintaining alignment, we start with separate models to learn from the 2D and 3D modalities. Inspired by the ``\textit{align before fuse}'' strategy in vision-language models \cite{albef,bao2022vlmo}, we introduce parameter sharing across the models, allowing our framework to learn a unified representation that integrates information from both modalities. Second, to enable flexible multi-modal learning with any combination of available modality data (i.e., molecules with only 2D or 3D information, as well as molecules with both 2D and 3D information), we employ 
2D$\rightarrow$3D and 3D$\rightarrow$2D decoders
to generate the missing modality, ensuring effectiveness even when only single-modality information is available.

\stitle{Summary of Contributions.}
Our proposed framework, FlexMol, supports flexible input from either single or paired modalities, as well as their mixture. Its effective feature alignment and fusion strategy  enables it to achieve competitive performance across various molecular property prediction tasks.  Trained on a dataset with only \emph{3.4M paired and 2M single-modality samples}, it can outperform much larger molecular models pre-trained on data \emph{exceeding 10M samples} on certain benchmark tasks. The main contribution of this paper is summarized as follows. 
\begin{itemize}[leftmargin=*]
    \item We propose FlexMol, a unified framework for molecule pre-training that effectively aligns and fuses 2D and 3D modalities, while preserving modality-specific information.
    
   % \item We introduce a novel approach combining separate models for 2D and 3D data with shared parameters, enabling efficient fusion of modality-specific representations.

    \item We develop 
    % a decoder mechanism 
2D$\rightarrow$3D and 3D$\rightarrow$2D decoders
    that can generate missing modality data based on the available modality, allowing the model to perform multi-modal learning even with single-modality inputs. Hence, FlexMol supports flexible pre-training data, including a mixture of paired and single-modality data.
    
    \item We empirically demonstrate that FlexMol achieves competitive performance across various benchmark tasks in molecular property prediction.
    % which could outperform caertain large-scale molecule pre-training methods in specific cases.
\end{itemize}

\section{Related Work}
\stitle{2D Molecule Pre-training.}
2D molecule pre-training focuses on learning molecular representations from graph-based structures, often incorporating graph augmentations or sequential SMILES representations.
PretrainGNN \cite{Hu_2020Strategies} is a pre-training strategy for graph neural networks (GNNs) that combines node- and graph-level tasks to capture both local and global representations. Building on similar ideas of leveraging multi-level information, GROVER \cite{NEURIPS2020_GROVER} employs self-supervised tasks at node, edge, and graph levels to capture structural and semantic information. Expanding the scope of molecular pre-training, MolCLR \cite{Wang2022} focuses on self-supervised learning with graph augmentations, pre-training on 10 million unlabelled molecules via atom masking, bond deletion, and subgraph removal. Complementary to this, DVMP \cite{dual_view_kdd23} introduces a dual-view framework by integrating Transformer and GNN branches to harness both sequential (SMILES) and graphical representations of molecules. Taking a further step in node- and graph-level learning, Mole-BERT \cite{xia2023molebert} employs a context-aware tokenizer using VQ-VAE, enabling masked atom modeling and triplet masked contrastive learning to refine molecular representations. FineMolTex \cite{yiboli2025kdd} enables the model not only to establish correspondences between entire molecular graphs and their textual descriptions but also to align common 2D motifs with key terms in descriptions, enhancing the understanding of molecular structures.

\stitle{3D Molecule Pre-training.}
3D molecule pre-training leverages geometric information such as atomic distances, bond angles, and 3D conformations to capture spatial and physical properties of molecules.
GEM \cite{Fang2022_gem} employs a geometry-based graph neural network with geometry-level self-supervised learning strategies to integrate bond angles and bond lengths as additional edge attributes, enhancing the representation of 3D molecular information. Building on this emphasis on 3D geometry, GeoSSL-DDM \cite{liu2023molecular} uses an SE(3)-invariant score matching strategy to denoise pairwise atomic distances while leveraging energy-based models (EBM) for mutual information maximization, offering a generative self-supervised learning method for molecular geometric data. Diverging from the pairwise focus of GeoSSL-DDM, LEGO \cite{LEGO3d} targets localized tetrahedral structures as core representations, employing perturbation and reconstruction of these units with masked modeling to pre-train molecular representations. Expanding the application of 3D data, Uni-Mol \cite{zhou2023unimol} introduces a unified framework with two SE(3) Transformer models pre-trained for molecular conformations and protein pockets. %Further innovating with noise-based strategies, 
Furthermore, Frad \cite{Ni2024_NMI} adopts fractional denoising and hybrid noise designs, incorporating chemical priors to enhance force learning interpretation and achieve refined molecular distribution modeling. 

\stitle{2D \& 3D Molecule Pre-training.}
%\paragraph{Separate modality training}
We first review separate modality training methods.
GraphMVP \cite{graphmvp} employs a multiview framework for molecular pre-training by maximizing mutual information (MI) between 2D and 3D representations, reformulating MI via conditional probabilities and leveraging contrastive and generative losses for robust integration. MoleculeSDE \cite{molsde} extends this with direct data-space modeling for geometry and topology reconstruction, utilizing SE(3)-equivariant and reflection-antisymmetric SDEs.
In contrast, 3D Infomax \cite{3dinfomax} encodes implicit 3D knowledge into GNNs using only 2D graphs, maximizing MI between latent 3D representations and GNN outputs. This allows models to infer 3D geometry during fine-tuning without explicit 3D data, capturing transferable 3D features for efficient 2D-based inference.
Unicorn \cite{feng2024unicorn} integrates 2D graph masking, 2D-3D contrastive learning, and 3D denoising via a diffusion process to model augmented trajectories. MoleculeJAE \cite{du2023molecule} further unifies 2D and 3D representations through self-supervised chemical structure learning.

%\paragraph{Unified modality training}
The other line of research focuses on unified modality training.
Zhu et al. \cite{unified2d3d_kdd} propose a unified 2D-3D molecular pre-training framework with three tasks: masked atom/coordinate reconstruction, 2D-to-3D conformation generation, and 3D-to-2D graph generation on a backbone graph network block \cite{addanki2021largescalegraphrepresentationlearning}.
To enhance multimodal integration, Transformer-M \cite{luo2023transformerunderstand2d} employs separate channels for 2D and 3D structures but unifies them in a Transformer-based model, enabling flexible processing of both formats. MolBlend \cite{yu2024multimodal} follows the same encoding approach yet adopts self-supervised pre-training, unifying 2D-3D molecular relations into a single matrix and reconstructing modality-specific information, thereby improving generalization.

\section{Proposed Method}
\begin{figure*}
    \centering
    \includegraphics[width=\linewidth]{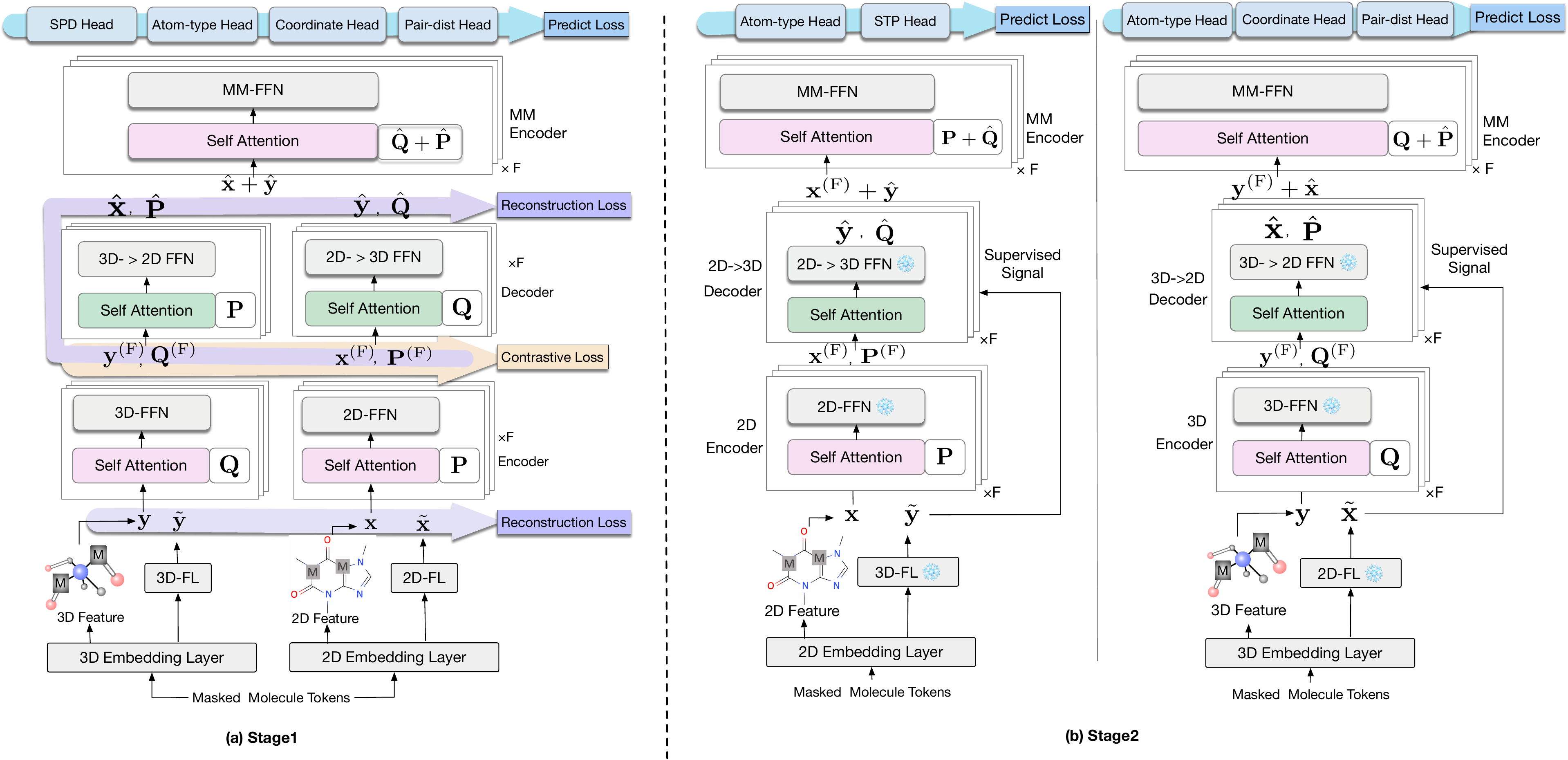}
    \vspace{-4mm}
    \caption{FlexMol framework pipeline. Stage 1: Pre-training unified molecular representation using paired 2D \& 3D modalities. Stage 2: Continuous training with single modality molecule data, where the left side represents the 2D-only scenario and the right side represents the 3D-only scenario. The self-attention blocks with the same color indicate shared parameters, while the snowflake icon represents frozen parameters.}
    \label{fig:pre-train-stage}
\end{figure*}

In the first pre-training stage, paired 2D and 3D molecular features are used for self-supervised training to learn a unified molecular representation. In the second pre-training stage, continual learning is conducted on the model trained in the first stage using single-modality data. The overall pipeline is illustrated in Figure \ref{fig:pre-train-stage}.

\subsection{Pre-training on Paired 2D \& 3D Modalities}

% \subsubsection{Architecture Overview}

\subsubsection{2D \& 3D Feature Learning}
Following prior work Transformer-M \cite{luo2023transformerunderstand2d}, we represent the atoms and their associated features as a matrix \(\mathbf{X} \in \mathbb{R}^{n \times d}\), where \(n\) denotes the number of atoms, and \(d\) is the dimensionality of the feature space.  
For a 2D molecular structure, we define the molecular graph as \(\mathcal{G}_{2D} = (\mathbf{X}, E)\), where \(E\) represents the set of edges. An edge \(e(i, j) \in E\) corresponds to the feature of the bond (e.g., bond type) between atom \(i\) and atom \(j\), provided such a bond exists.  
For the 3D geometric structure, we define it as \(\mathcal{G}_{3D} = (\mathbf{X}, R)\), where \(R = \{r_1, r_2, \dots, r_n\}\) is a set of 3D coordinates, with each \(r_i \in \mathbb{R}^3\) specifying the spatial position of atom \(i\).

\stitle{2D Molecule Feature.}
For each atom \(i\), let \(\psi_{\text{deg}}(i)\) represent the degree encoding of atom \(i\), which is a \(d\)-dimensional learnable vector determined by the atom's degree. The degree encodings for all atoms in the molecule are collectively denoted as \(\Psi_{\text{2D}} = [\psi_{\text{deg}}(1), \psi_{\text{deg}}(2), \ldots, \psi_{\text{deg}}(n)] \in \mathbb{R}^{n \times d}\).
The representation of a 2D molecule is then given by  
$
\mathbf{x} = \mathbf{X} + \Psi_{\text{2D}}, \text{ where } \mathbf{x} \in \mathbb{R}^{n \times d},
$
and \(\mathbf{X} \in \mathbb{R}^{n \times d}\) denotes the initial features of the atoms.  

\stitle{3D Molecule Feature.}
For each atom pair $(i,j)$, we compute a distance encoding $\psi(i,j) \in \mathbb{R}^K$, where each component $\psi^k(i,j)$ is obtained by applying a Gaussian kernel $k$ to the Euclidean distance between $i$ and $j$, where $k = 1, \ldots, K$, where $K$ is the number of Gaussian Basis kernels, following Transformer-M \cite{luo2023transformerunderstand2d}.
For each atom \(i\), the 3D distance encodings between \(i\) and all other atoms are summed to compute its centrality encoding:  
$\psi_{\text{3D}}(i) = \sum_{j=1}^n \psi(i, j) \mathbf{W}_D,$ 
where \(\mathbf{W}_D \in \mathbb{R}^{K \times d}\) is a learnable weight matrix, and \(\psi(i, j) \in \mathbb{R}^K\) aggregates the Gaussian kernel values for each pair.  
The 3D molecule representation is then given by  
$\mathbf{y} = \mathbf{X} + \Psi_{\text{3D}}, \text{ where } \mathbf{y} \in \mathbb{R}^{n \times d},$ 
and \(\Psi_{\text{3D}} = [\psi_{\text{3D}}(1), \ldots, \psi_{\text{3D}}(n)] \in \mathbb{R}^{n \times d}\).  

\stitle{2D \& 3D Molecule Representation Learning.}  
We employ two separate multi-layer perceptrons (MLPs) to approximate the previously obtained 2D and 3D molecular features, \( \mathbf{x} \) and \( \mathbf{y} \). The outputs of these MLPs are denoted as \( \tilde{\mathbf{x}} \) and \( \tilde{\mathbf{y}} \), which serve as the supervision signals for the decoders of the missing modality during the second stage of training.

\stitle{2D Atom Pair Representation.} 
% For each atom pair \((i,j)\), let \(d_{ij} \in \mathbb{Z}_{\ge 0}\) be the Shortest-Path Distance in the molecular graph. 
% We denote $\Phi^{\mathrm{SPD}}_{ij} \in \mathbb{R}$ as a encoding between atom $i$ and $j$, which is a learnable scalar determined by $d_{ij}$.
% The edge encoding is defined as $\Phi_{i j}^{\text {Edge}}=\frac{1}{N} \sum_{n=1}^N \mathbf{e}_n\left(w_n\right)^T$, where \(w_n\) are learnable vectors matching the edge feature dimension. Then, the 2D atom pair representation is $\mathbf{P} = \Phi^{\text {SPD}}+\Phi^{\text {Edge }} \in \mathbb{R}^{n \times n}$.
For each atom pair $(i,j)$, let $d_{ij} \in \mathbb{Z}_{\ge 0}$ be the Shortest-Path Distance (SPD) in the molecular graph.  
We denote $\Phi^{\mathrm{SPD}}_{ij} \in \mathbb{R}$ as an SPD encoding between atom $i$ and $j$, which is a learnable scalar determined by the distance of the shortest path between them.  
In addition, we encode the edge features (e.g., chemical bond types) along the shortest path from $i$ to $j$.  
Let the sequence of edges on this shortest path be $STP_{ij} = (e_1, e_2, \ldots, e_N)$, where each $e_n$ denotes the feature vector of the $n$-th edge on the path.  
For most molecules, there exists only one distinct shortest path between any two atoms; in the rare case of multiple shortest paths, we simply take one returned by the shortest-path algorithm.  
The edge encoding between $i$ and $j$ is defined as  
$\Phi_{ij}^{\text{Edge}} = \frac{1}{N} \sum_{n=1}^N e_n (w_n)^T$,
where $w_n$ are learnable vectors of the same dimension as the edge features.  
Finally, the 2D atom pair representation is obtained as  $\mathbf{P} = \Phi^{\text{SPD}} + \Phi^{\text{Edge}} \in \mathbb{R}^{n \times n}$.

\stitle{3D Atom Pair Representation.} The 3D distance encoding $\Phi^{\text {3D}}_{ij}$ is obtained according to
$\Phi^{\text {3D}}_{ij}=\operatorname{GELU}\left(\boldsymbol{\psi}_{(i, j)} \boldsymbol{W}_D^1\right) \boldsymbol{W}_D^2$, where $\boldsymbol{\psi}_{(i, j)}=\left[\psi_{(i, j)}^1 ; \ldots ; \psi_{(i, j)}^K\right]^{\top}$ is the Gaussian basis kernel applied to the $d_{ij}$ capturing spatial variations. $\boldsymbol{W}_D^1 \in \mathbb{R}^{K \times K}, \text{ and } \boldsymbol{W}_D^2 \in \mathbb{R}^{K \times 1}$are learnable parameters.
Subsequently, the 3D atom pair representation is $\mathbf{Q} = \Phi^{\text {3D}} \in \mathbb{R}^{n \times n}.$

\subsubsection{Transformer layers}  
Inspired by some vision-language models \cite{albef,bao2022vlmo}, we adopt an ``align before fuse'' Transformer architecture to jointly learn 2D and 3D molecular representations, which enables efficient encoding of both features. The 2D and 3D representations are further aligned using contrastive learning, as detailed in \S\ref{sec:losses}.

\stitle{2D (3D) Encoder.}  
We adopt the SE(3) Transformer encoder with \( F \) layers, as proposed in Uni-Mol \cite{zhou2023unimol}, to encode molecular structures with 3D equivariance. To process 2D and 3D modalities efficiently, the encoder shares self-attention parameters.  
The encoder inputs 2D/3D molecular representations (\( \mathbf{x}, \mathbf{y} \)) and atom pair representations (\( \mathbf{P}, \mathbf{Q} \)) as attention bias. After \( F \) layers, it outputs \( \mathbf{x}^{(F)}, \mathbf{y}^{(F)} \) (molecules) and \( \mathbf{P}^{(F)}, \mathbf{Q}^{(F)} \) (pairs).  

We maintain a pair-level representation, similar to Uni-Mol \cite{zhou2023unimol}. Take 2D atom pair representation as example, \( \mathbf{P}_{ij} \) is initialized as the 2D atom pair representation and iteratively updated via atom-to-pair communication using multi-head Query-Key interaction. The update for atom pair \( ij \) at layer \( l+1 \) is:  
\begin{equation}
    \mathbf{P}_{ij}^{(l+1)} = \mathbf{P}_{ij}^{(l)} + \left\{ \left. \textstyle\frac{\mathbf{Z}_i^{(l, h)} (\mathbf{K}_j^{(l, h)})^\top}{\sqrt{d}} \right| h \in \{1, \ldots, H\} \right\},
\end{equation}  
where \( H \) is the number of attention heads, \( d \) is the hidden dimension, and \( \mathbf{Z}_i^{(l, h)} \), \( \mathbf{K}_j^{(l, h)} \) denote Query and Key of atom \( i, j \) in the \( h \)-th attention head. Similarly, \( \mathbf{Q}_{ij} \) is updated.

\stitle{2D$\rightarrow$3D (3D$\rightarrow$2D) Decoder.}  
We further propose an $F$-layer SE(3)-Transformer decoder to reconstruct missing modality features. The decoder uses cross-attention to integrate the aligned 2D and 3D representations while ensuring 3D equivariance.  

The 3D$\rightarrow$2D decoder inputs the 3D molecular representation \( \mathbf{y}^{(F)} \) and uses \( \mathbf{P} \) as self-attention bias. Similarly, the 2D$\rightarrow$3D decoder inputs \( \mathbf{x}^{(F)} \) and uses \( \mathbf{Q} \) as self-attention bias. The self-attention layers share parameters in both decoders.  

We utilize cross-attention to align the two modalities. In the 3D$\rightarrow$2D decoder, the cross-attention computes:  
\begin{equation}
\text{Attention}_{2D}(\mathbf{y}^{(F)}, \mathbf{x}^{(F)}) = \text{softmax}\left(\textstyle\frac{\mathbf{y}^{(F)} \mathbf{x}^{(F)\top}}{\sqrt{d}}\right) \mathbf{x}^{(F)},
\end{equation}  
where \( \sqrt{d} \) is a scaling factor based on the dimensionality of the embeddings.
Conversely, in the 2D$\rightarrow$3D decoder, cross-attention uses \( \mathbf{x}^{(F)} \) as the query and \( \mathbf{y}^{(F)} \) as the key and value.  

The 3D$\rightarrow$2D decoder outputs \( \hat{\mathbf{x}} \) and \( \hat{\mathbf{P}} \), while the 2D$\rightarrow$3D decoder outputs \( \hat{\mathbf{y}} \) and \( \hat{\mathbf{Q}} \):  
\begin{align}
\begin{split}
        &\hat{\mathbf{x}}, \hat{\mathbf{P}} = \text{Decoder}_{2D}(\mathbf{y}^{(F)}, \mathbf{Q}, \mathbf{x}^{(F)}, \mathbf{P}),\\
    &\hat{\mathbf{y}}, \hat{\mathbf{Q}} = \text{Decoder}_{3D}(\mathbf{x}^{(F)}, \mathbf{P}, \mathbf{y}^{(F)}, \mathbf{Q}).
\end{split}
\end{align}  
The decoders integrate complementary information and reconstruct the representations.  
We then use a reconstruction loss (\S\ref{sec:losses}) to further guide the decoder, ensuring modality consistency and robustness to missing data.

\stitle{Multi-modal Encoder.}  
The multi-modal (MM) encoder refines the approximate molecular representations \(\hat{\mathbf{x}}\) and \(\hat{\mathbf{y}}\) from the decoders and utilizes the decoder-learned atom-pair features \(\hat{\mathbf{P}}\) and \(\hat{\mathbf{Q}}\) as attention bias. Similar to the vision-language expert in models like VLMo \cite{bao2022vlmo}, the encoder integrates cross-modal interactions while preserving intra-modal information.
After \( L \) layers, the refined representations \(\mathbf{x}^{(L)}\) and \(\mathbf{y}^{(L)}\) are obtained as the final outputs of the encoder. These representations are used as input to various 2D and 3D prediction heads for downstream tasks, enabling effective multi-task learning.

\subsubsection{Pre-training Target}\label{sec:losses}

Our pre-training employs several different losses, as follows.

\stitle{Contrastive Loss.}
We use InfoNCE loss to align the 2D \& 3D representations generated by the 2D and 3D encoders.

\begin{align}
\mathcal{L}_{cl} 
&= -\frac{1}{2} \mathbb{E}_{p(\boldsymbol{x}, \boldsymbol{y})} \Bigg[ 
    \log \textstyle\frac{\exp \left(\left\langle \mathbf{x}^{(F)}, \mathbf{y}^{(F)} \right\rangle \right)}{
        \exp \left(\left\langle \mathbf{x}^{(F)}, \mathbf{y}^{(F)} \right\rangle \right) 
        + \sum_j \exp \left(\left\langle \mathbf{x}^{(F)}_{j}, \mathbf{y}^{(F)} \right\rangle \right)} \notag \\
&\quad +
    \log \textstyle \frac{\exp \left(\left\langle \mathbf{y}^{(F)}, \mathbf{x}^{(F)} \right\rangle \right)}{
        \exp \left(\left\langle \mathbf{y}^{(F)}, \mathbf{x}^{(F)} \right\rangle \right) 
        + \sum_j \exp \left(\left\langle \mathbf{y}^{(F)}_{j}, \mathbf{x}^{(F)} \right\rangle \right)} 
\Bigg].
\end{align}

\stitle{Reconstruction Loss.} The reconstruction loss consists of two components, as follows.
\paragraph{Representation Alignment Loss:} This loss measures the discrepancy between the molecular representations learned by the modality-specific Feature Learner (FL), namely, 2D/3D-FL (i.e., MLPs in our framework), and their corresponding reconstructed counterparts. For the 2D modality, it aligns the 2D molecular representation \(\mathbf{x}\) with the representation learned by the 2D-FL, \(\tilde{\mathbf{x}}\). Similarly, for the 3D modality, it aligns the 3D molecular representation \(\mathbf{y}\) with the representation learned by the 3D-FL, \(\tilde{\mathbf{y}}\), as follows.
\begin{equation}
    \mathcal{L}_{\text{ra}} = \|\mathbf{x} - \tilde{\mathbf{x}}\|^2 + \|\mathbf{y} - \tilde{\mathbf{y}}\|^2.
\end{equation}

\paragraph{Encoder-Decoder Consistency Loss:} This loss ensures consistency between the multi-layer encoder's learned molecular and atom pair representations and the outputs generated by the decoder. Specifically, the 2D/3D molecular representations \(\mathbf{x}^{(F)}\) and \(\mathbf{y}^{(F)}\) learned by the encoder are aligned with the reconstructed molecular representations \(\hat{\mathbf{x}}\) and \(\hat{\mathbf{y}}\) produced by the decoder. Similarly, the 2D/3D atom pair representations \(\mathbf{P}\) and \(\mathbf{Q}\) learned by the encoder are aligned with their reconstructed counterparts \(\hat{\mathbf{P}}\) and \(\hat{\mathbf{Q}}\), as follows.
\begin{equation}
    \mathcal{L}_{\text{c}} =  \|\mathbf{x}^{(F)} - \hat{\mathbf{x}}\|^2 + \|\mathbf{y}^{(F)} - \hat{\mathbf{y}}\|^2 + \|\mathbf{P} - \hat{\mathbf{P}}\|^2 + \|\mathbf{Q} - \hat{\mathbf{Q}}\|^2.
\end{equation}
The overall reconstruction loss is the sum of these two components:
$\mathcal{L}_{\text{rec}} = \mathcal{L}_{\text{ra}} + \mathcal{L}_{\text{c}}.$

\stitle{Prediction Head.} 
%Following Uni-Mol \cite{zhou2023unimol}, we use the same 3D position recovery and masked atom prediction tasks as the pre-training target. %Additionally, we introduce a 2D Shortest Path Prediction (STP) task. 
% TODO: introduce stp head
%Using the STP head in Figure \ref{fig:pre-train-stage}(a), we aim to enhance the capability of the model to capture the 2D graph-level structural information.
Following Uni-Mol \cite{zhou2023unimol}, we adopt the same pre-training objectives of 3D position recovery and masked atom prediction. Additionally, we introduce a shortest path distance (SPD) prediction task, implemented with a two-layer MLP head, to incorporate 2D molecular graph features as auxiliary self-supervised signals alongside Uni-Mol's 3D-based objectives.

\subsection{Pre-training in Single Modality}

The second pre-training stage pipeline is shown in Figure \ref{fig:pre-train-stage}(b). In this stage, the model is complemented with only a single modality (either 2D or 3D molecular data), which fine-tunes the representations learned in the first stage.  

In the 2D-only scenario, we first utilize the frozen 3D-FL, trained during the first stage, to generate the 3D molecular representation \(\tilde{\mathbf{y}}\). This serves as the supervision signal for 2D$\rightarrow$3D decoder, which learns to generate the 3D molecular representation \(\hat{\mathbf{y}}\) and the corresponding atom pair representation \(\hat{\mathbf{Q}}\).  

Next, the 2D molecular representation \(\mathbf{x}^{(F)}\), output by the 2D encoder, is combined with the decoder-generated 3D molecular representation \(\hat{\mathbf{y}}\). The resulting fused representation is then passed as input to the multi-modal encoder. Additionally, the 3D atom pair representation \(\hat{\mathbf{Q}}\), generated by the decoder, is combined with the original 2D atom pair representation \(\mathbf{P}\) to construct the attention bias. Specifically, the pairwise term \(\hat{\mathbf{Q}} + \mathbf{P}\) is added to the attention score before applying the softmax function, allowing the model to incorporate structural information from both modalities during the self-attention computation.  

This process enables the multi-modal encoder to effectively combine 2D and 3D molecular features, achieving a comprehensive and robust molecular representation.  

\section{Experiment}
We conduct experiments on molecular property prediction and conformation generation, and analyze the model’s performance.

\subsection{Molecular Property Prediction}

\subsubsection{Experimental Setup}
All experiments were conducted on a server with an Intel Xeon Gold 2.40 GHz CPU and NVIDIA A100 40GB GPUs. For the pre-training Stage 1, we used PyTorch Lightning for distributed training on 2 GPUs for 20 epochs. Each epoch takes 3.5 hours, resulting in a total of 140 GPU hours. For the pre-training Stage 2, our model takes 4 GPU hours on 3D-only molecule data and 7 GPU hours on 2D-only molecule data for 10 epochs. 

\stitle{Datasets.} We train FlexMol using the aforementioned two-stage pre-training approach. 

In Stage 1 of the pre-training, we utilize \textbf{PCQM4Mv2}, a dataset containing paired 2D and 3D information \cite{pcqmv2}. It comprises 3.4 million organic molecules sourced from PubChemQC, each with a single equilibrium conformation and a label derived from DFT calculations. Since our approach is self-supervised, the label is not used. Each molecule is represented as: (a) a 2D graph with nodes as atoms and edges as chemical bonds; (b) a SMILES string, which can generate graph representations; (c) 3D structural information (coordinates) to enhance model performance.

During the second pre-training stage, we used the dataset provided by Uni-Mol \cite{zhou2023unimol}. The
Uni-Mol dataset is 3D-feature data, which contains about 19M molecules, and uses RDKit to randomly generate 11 conformations for each molecule. To construct our training set for Stage 2, we extracted a subset of 2 million molecules, referred to as \textbf{Unimol-2M-3D}, which retains the 3D structural information, including atomic coordinates and spatial relationships such as atom pair distances. 
Additionally, for each molecule in this 2M subset, we used its SMILES representation to generate corresponding 2D molecular features, forming the \textbf{Unimol-2M-2D} dataset, which contains 2D features like edge\_index, edge\_input, spatial\_pos, and in\_degree.

\stitle{Baselines.}
We use baseline models trained on datasets of varying sizes, as follows.
\begin{itemize}[leftmargin=*]
\item Supervised methods such as Attentive FP, and N-Gram\textsubscript{RF}, along with the unsupervised method trained on MoleculeNet \cite{Wu2018_molenet}. 

\item PretrainGNN \cite{Hu_2020Strategies} and Mole-BERT \cite{xia2023molebert}, which use 2D molecular features and are pre-trained on 2M samples from ZINC15 \cite{zinc15}. 

\item 3D InfoMax \cite{3dinfomax}, GraphMVP \cite{graphmvp}, MoleculeSDE \cite{molsde}, MoleBlend \cite{yu2024multimodal}, and Transformer-M \cite{luo2023transformerunderstand2d} utilize both 2D and 3D modalities to pre-train molecular representations on 3.4M samples from the PCQM4Mv2 \cite{pcqmv2} or other small-scale paired 2D-3D datasets. 

\item GROVER \cite{NEURIPS2020_GROVER}, MolCLR \cite{Wang2022}, GEM \cite{Fang2022_gem}, and Uni-Mol \cite{zhou2023unimol} are 3D molecular models pre-trained on datasets containing over 10M 3D samples. 
\end{itemize}

\begin{table*}[tb]
\centering
\caption{Performance (ROC-AUC \%) on molecular property (2D topology) classification tasks. Results are averaged over 3 runs with standard deviations in parentheses. Results of models with $[\diamond]$ are taken from Uni-Mol \cite{zhou2023unimol} and $[\dagger]$ from MoleBlend \cite{yu2024multimodal}.
Transformer-M* denotes our reproduced variant of Transformer-M, differing only in its training objective: the original model was optimized with PCQM labels, whereas we adopt purely self-supervised signals.
}
\label{tab:uni-mol-performance}
\begin{tabular}{l|c|c|ccccccc}
\hline
%& & &  \multicolumn{7}{c}{\textbf{ROC-AUC \%} ($\uparrow$)} \\
\textbf{Dataset} & \textbf{Pre-train size} & \textbf{Modality} & \textbf{BBBP} & \textbf{BACE}  & \textbf{Tox21} & \textbf{ToxCast} & \textbf{SIDER} & \textbf{HIV} & \textbf{PCBA} \\ \hline
\# Molecules & & & 2039 & 1513  & 7831 & 8575 & 1427 & 41127 & 437929  \\ 
\# Tasks & & & 1 & 1  & 12 & 617 & 27 & 1 & 128 \\ 
\hline
\multicolumn{10}{c}{Models trained on small or similar-scale data } \\
\hline
% D-MPNN $[\diamond]$           & -    & 2D & 71.0 (0.3) & 80.9 (0.6) & 75.9 (0.7) & 65.5 (0.3) & 57.0 (0.7) & 77.1 (0.5) & 86.2 (0.1) \\
Attentive FP  $[\diamond]$      & -    & 2D & 64.3 (1.8) & 78.4 (0.022) & 76.1 (0.5) & 63.7 (0.2) & 60.6 (3.2) & 75.7 (1.4) & 80.1 (1.4) \\
N-Gram\textsubscript{RF}  $[\diamond]$& -    & 2D & 69.7 (0.6) & 77.9 (1.5)  & 74.3 (0.4) & 66.2 (0.5) & 66.8 (0.7) & 77.2 (0.1) & 75.0 (0.2) \\
PretrainGNN $[\diamond]$        & 2M   & 2D & 68.7 (1.3) & 84.5 (0.7) & 78.1 (0.6) & 65.7 (0.6) & 62.7 (0.8) & 79.9 (0.7) & 86.0 (0.1) \\
3D InfoMax $[\dagger]$         & 1.1M & Mixed  & 70.4 (1.0) & 79.7 (1.5) & 74.5 (0.7) & 64.4 (0.8) & 60.6 (0.7) & 76.1 (1.3) & 75.2 (0.4) \\
GraphMVP $[\dagger]$           & 50K  & Mixed  & 68.5 (0.2) & 76.8 (1.1) & 72.5 (0.4) & 62.7 (0.1) & 62.3 (1.6) & 74.5 (0.5) & 73.0 (0.3) \\
MoleculeSDE $[\dagger]$        & 3.4M & Mixed  & 71.8 (0.7) & 79.5 (2.1) & 76.8 (0.3) & 65.0 (0.2) & 60.8 (0.3) & 78.8 (0.9) & 74.8 (0.6) \\
Mole-BERT $[\dagger]$         & 2M   & 2D & 71.9 (1.6) & 80.8 (1.4) & 76.8 (0.5) & 64.3 (0.2) & 62.8 (1.1) & 78.2 (0.8) & 76.0 (0.4) \\
MoleBlend $[\dagger]$          & 3.4M & Mixed  & {73.0}(0.8) & 83.7 (1.4) & 77.8 (0.8) & 66.1 (0.0) & 64.9 (0.3) & \textbf{79.0} (0.8) & 75.5 (0.5) \\
Transformer-M* &3.4M & Mixed & 69.7 (0.6) & 78.1 (0.7) & 77.4 (0.4) & 62.6 (0.2) &62.1 (0.3) & 76.3 (0.2) & 86.1 (0.3) \\
Uni-Mol            & 3.4M & 3D & 66.0 (0.7) & 75.8 (0.6) & 72.0 (0.5) & 61.3 (0.3) & 58.8 (0.4) & 74.0 (0.6) & 83.5 (0.3) \\
Uni-Mol            & 5.4M & 3D & 69.2 (0.6) & 77.9 (0.5) & 74.1 (0.4) & 62.7 (0.3) & 60.5 (0.3) & 74.9 (0.5) & 84.7 (0.3) \\
\hline
\multicolumn{10}{c}{Our models} \\
\hline
FlexMol$^+$\textsubscript{2D}         & 5.4M & Mixed  & 72.4 (0.5) & 80.0 (0.3) & 77.6 (0.4) & 64.2 (0.3) & 63.0 (0.2) & 75.3 (0.4) & 86.0 (0.2) \\
FlexMol$^+$\textsubscript{3D}          & 5.4M & Mixed  & \textbf{75.1} (0.6) & \textbf{85.7} (0.5) & \textbf{78.6} (0.4) & \textbf{66.4} (0.3) & \textbf{65.3} (0.2) & 78.3 (0.3) & \textbf{86.6} (0.2) \\
\hline
\rowcolor{gray!20}
\multicolumn{10}{c}{Models trained on over 10M data (for reference only)} \\
\hline
\rowcolor{gray!20}
GROVER\textsubscript{base} $[\diamond]$  & 11M & 3D & 70.0 (0.1) & 82.6 (0.7) & 74.3 (0.1) & 65.4 (0.4) & 64.8 (0.6) & 62.5 (0.9) & 76.5 (2.1) \\
\rowcolor{gray!20}
GROVER\textsubscript{large} $[\diamond]$ & 11M & 3D & 69.5 (0.1) & 81.0 (1.4) & 73.5 (0.1) & 65.3 (0.5) & 65.4 (0.1) & 68.2 (1.1) & 83.0 (0.4) \\
\rowcolor{gray!20}
MolCLR  $[\diamond]$            & 10M  & 3D & 72.2 (2.1) & 82.4 (0.9) & 75.0 (0.2) & 66.5 (0.7) & 58.9 (0.1) & 78.1 (0.5) & 74.7 (0.3) \\
\rowcolor{gray!20}
GEM $[\diamond]$ & 20M & 3D & 72.4 (0.4) & 85.6 (1.1)  & 78.1 (0.1) & 69.2 (0.4) & 67.2 (0.4) & 80.6 (0.9) & 86.6 (0.1) \\ 
\rowcolor{gray!20}
Uni-Mol $[\diamond]$ & 19M & 3D &72.9 (0.6) & 85.7 (0.2) & 79.6 (0.5) & 69.6 (0.1) & 65.9 (1.3) & 80.8 (0.3) & 88.5 (0.1)  \\ 

% DVMP & 10M & 77.8 ± 0.3 & 89.4 ± 0.8 & 79.1 ± 0.4 & - & 69.8 ± 0.6 & 81.4 ± 0.4 & -\\ 

\hline
\end{tabular}
\end{table*}

\stitle{Hyperparameter Settings.}
The key hyperparameters and training settings of our model are listed below. 

During pre-training, the Adam optimizer is applied with default betas, a learning rate of \(3\cdot 10^{-5}\), and  30 epochs for the two stages in total. Transformer encoder/decoder layers are searched from \(F \in \{4, 6, 8\}\). The dimension of both the encoder and decoder is fixed at 512, and the number of attention heads is set to 64.

During fine-tuning for downstream tasks, we perform hyperparameter search for learning rate in \(\{10^{-5}, 3\cdot 10^{-5}, 10^{-4}\}\), batch size in \{16, 32, 64, 128\}, and dropout rate in \{0.1, 0.2, 0.3\}. Additionally, we use the `use\_lora' hyperparameter to control whether LoRA (Low-Rank Adaptation) \cite{hu2021loralowrankadaptationlarge} is applied during fine-tuning. When `use\_lora' is enabled, the `lora\_rank' is set to 64, allowing for efficient fine-tuning of the model while reducing the number of trainable parameters.

\subsubsection{Main Results}
Following MoleculeNet \cite{Wu2018_molenet}, we adopt scaffold splitting and report results for 2D- and 3D-based molecular property prediction in Tables~\ref{tab:uni-mol-performance} and \ref{tab:uni-mol-regression}, respectively. 
%Here \#N\_Mol is the pre-training data size, Mod denotes 2D/3D or mixed modality (M), 
Here FlexMol$^+$\textsubscript{2D (3D)} indicates pre-training on 3.4M paired data (PCQM4Mv2) followed by 2M 2D (3D)-only data. Baselines with smaller or similar-scale datasets are compared, with best results in bold; models trained with $>$10M molecules are for reference only.

Results across various datasets reveal that FlexMol$^+$\textsubscript{3D (2D)} consistently outperforms the baseline models with smaller or similar-scale datasets, while also achieving competitive or even superior performance compared to large-scale state-of-the-art baselines such as GEM and Uni-Mol in many cases. 

We further make two noteworthy observations.
(1) The performance of Uni-Mol drops significantly when trained on smaller data, suggesting its dependence on large-scale pre-training. Our model performs well even with limited data and consistently surpasses Uni-Mol at the same scale (5.4M), demonstrating that gains come from the 2D features and modality alignment.
(2) %Transformer-M* denotes our reproduced variant of Transformer-M, differing only in its training objective: the original model was optimized with PCQM labels, whereas we adopt purely self-supervised signals.  
Compared to Transformer-M* with identical feature construction, FlexMol's stronger performance implies that our Stage 2 pre-training with added single-modal data yields significant gains.

\begin{table*}[tbp]
\centering
\caption{Performance (RMSE \& MAE) on molecular property (3D conformation) regression tasks. Notations follow Table~\ref{tab:uni-mol-performance}.}
\label{tab:uni-mol-regression}
\small
\begin{tabular}{l|c|c|ccc|ccc}
\hline
& & & \multicolumn{3}{c|}{\textbf{RMSE} $\downarrow$} & \multicolumn{3}{c}{\textbf{MAE} $\downarrow$} \\
\textbf{Datasets} & \textbf{Pre-train size} & \textbf{Modality} & \textbf{ESOL} & \textbf{FreeSolv} & \textbf{Lipo} & \textbf{QM7} & \textbf{QM8} & \textbf{QM9} \\ 
\hline
\# Molecules & & &1128 & 642 & 4200 & 6830 & 21786 & 133885 \\ 
\# Tasks & & &1 & 1 & 1 & 1 & 12 & 3 \\ 
\hline
\multicolumn{9}{c}{Models trained on small or similar-scale data} \\
\hline
% D-MPNN $[\diamond]$ & - &2D & 1.050 (0.008) & 2.082 (0.082) & 0.683 (0.016) & 103.5 (8.6) & 0.0190 (0.0001) & 0.00814 (0.00001) \\ 
Attentive FP $[\diamond]$ & - & 2D&  0.877 (0.029) & 2.073 (0.183) & 0.721 (0.0010) & 72.0 (2.7) & 0.0179 (0.001) & 0.00812 (0.00001) \\ 
N-Gram\textsubscript{RF} $[\diamond]$ & - & 2D & 1.074 (0.107) & 2.688 (0.085) & 0.812 (0.028) & 92.8 (4.0) & 0.0236 (0.0006) & 0.01037 (0.00016) \\ 
% N-Gram\textsubscript{XGB} & 1.083 (0.082) & 5.061 (0.744) & 2.072 (0.030) & 81.9 (1.9) & 0.0215 (0.0005) & 0.00964 (0.00031) \\ 
PretrainGNN $[\diamond]$ & 2M &2D & 1.100 (0.006) & 2.764 (0.002) & 0.739 (0.003) & 113.2 (0.6) & 0.0200 (0.0001) & 0.00922 (0.00004) \\ 
GraphMVP $[\diamond]$ & 50K & Mixed & 1.029 (0.033) & - & 0.681 (0.010) & - & 0.0178 (0.0003) & - \\ 
Transformer-M* & 3.4M & Mixed & 0.925 (0.034) &1.772 (0.058) & 0.723 (0.026) &  61.23 (1.3) & 0.0177 (0.0004) & 0.00608 (0.0004)\\
Uni-Mol & 3.4M & 3D &  0.959 (0.030) &
2.509 (0.052) &
0.774 (0.031)&
60.60 (0.2)&
0.0186 (0.0002)&
0.00649 (0.0004)\\
Uni-Mol & 5.4M & 3D &  0.912 (0.029) &
2.101 (0.055) &
0.745 (0.027)&
56.20 (0.3)&
0.0181 (0.0003)&
0.00612 (0.0004) \\
\hline
\multicolumn{9}{c}{Our models} \\
\hline
FlexMol$^+$\textsubscript{2D} & 5.4M & Mixed & 
0.918 (0.031) &
\textbf{1.623} (0.064) &
0.709 (0.020) &
\textbf{52.8} (1.5) & 0.0176 (0.0004) & 0.00589 (0.0003) \\
FlexMol$^+$\textsubscript{3D} &5.4M & Mixed & \textbf{0.812} (0.040) &1.738 (0.087) & \textbf{0.640} (0.028)
& 53.1 (2.2) &
\textbf{0.0170} (0.0005)&
\textbf{0.00561} (0.0004)\\
% FlexMol-M & + 1M 2D + 1M 3D & 
% 0.848 & 1.626 &  
% 0.702 &
% 55.0 &
% 0.0171 &
% 0.00590\\
\hline
\rowcolor{gray!20}
\multicolumn{9}{c}{Models trained on over 10M data (for reference only) } \\
 \hline
\rowcolor{gray!20}
GROVER\textsubscript{base} $[\diamond]$ & 11M & 3D& 0.983 (0.090) & 2.176 (0.052) & 0.817 (0.008) & 94.5 (3.8) & 0.0218 (0.0004) & 0.00984 (0.00055) \\
\rowcolor{gray!20}
GROVER\textsubscript{large} $[\diamond]$ & 11M & 3D & 0.895 (0.017) & 2.272 (0.051) & 0.823 (0.010) & 92.0 (0.9) & 0.0224 (0.0003) & 0.00986 (0.00025) \\ 
\rowcolor{gray!20}
MolCLR  $[\diamond]$& 10M & 3D & 1.271 (0.040) & 2.594 (0.249) & 0.691 (0.004) & 66.8 (2.3) & - & 0.00746 (0.00001) \\ 
\rowcolor{gray!20}
GEM $[\diamond]$ & 20M & 3D & {0.798} (0.029) & 1.870 (0.094) & 0.660 (0.008) & 58.9 (0.8) & 0.0171 (0.0001) & 0.00746 (0.00001) \\ 
\rowcolor{gray!20}
Uni-Mol $[\diamond]$ & 19M & 3D & 0.788 (0.029)& {1.480 (0.048)} & {0.603 (0.010)} & {41.8} (0.2) & {0.0156 (0.0001)} & {0.00467} (0.00004) \\ 
% DVMP & 10M & 0.817 ± 0.024 & 1.952 ± 0.061 & 0.653 ± 0.002 & 74.4 ± 1.2 & 0.0171 ± 0.0004 & -\\
\hline
\end{tabular}
\end{table*}

%Although Uni-Mol achieves exceptional scores on large-scale datasets like QM9, FlexMol$^+$\textsubscript{3D} balances scalability and accuracy, making it suitable for applications where both 2D and 3D features play critical roles.

\subsection{Molecule Conformation Generation}  
\subsubsection{Experimental Setup} 
Following prior work \cite{pmlr_gen,zhou2023unimol}, we evaluate our model on the conformation generation task using the GEOM-QM9 dataset \cite{goem}, which involves generating accurate and diverse 3D molecular conformations from corresponding 2D molecular graphs. Unlike traditional approaches that rely on expensive methods such as advanced sampling or semi-empirical DFT, recent methods \cite{gen1_Simm2020,gen2_Xu2020,gen3_Mansimov2019,gen4_Ganea2021} leverage learned representations for efficient conformation generation.

\stitle{Baselines.}
Following Uni-Mol, we select ten baseline methods. RDKit \cite{riniker2015better} is a classical conformation generation approach grounded in distance geometry. GraphDG \cite{simm2020generative}, CGCF \cite{xu2020learning}, ConfVAE \cite{xu2021end}, ConfGF \cite{shi2021learning}, and DGSM \cite{92} employ generative modeling in conjunction with distance geometry. These methods typically generate interatomic distance matrices as an intermediate representation, followed by the iterative reconstruction of atomic coordinates. CVGAE \cite{mansimov2019molecular}, GeoMol \cite{ganea2021geomol}, DMCG \cite{93}, and GeoDiff \cite{94} directly predict atomic coordinates, thereby circumventing the need for intermediate distance-based representations.

\stitle{Evaluation Metrics.}
We follow Uni-Mol \cite{zhou2023unimol} in using RDKit to generate initial conformations, and fine-tune our model to map 2D graphs to labeled 3D conformations. For each molecule, we generate twice the number of labeled conformations and select the closest prediction to each labeled conformation based on root-mean-square deviation (RMSD). Performance is evaluated using Coverage (COV) and Matching (MAT), where higher COV indicates greater diversity, and lower MAT reflects higher accuracy.
\begin{equation}
    \operatorname{COV}(S_g, S_r) = \textstyle\frac{\left|\left\{ \boldsymbol{R} \in S_r \mid \exists \hat{\boldsymbol{R}} \in S_g, \operatorname{RMSD}(\boldsymbol{R}, \hat{\boldsymbol{R}}) < \delta \right\}\right|}{|S_r|},
\end{equation}
\begin{equation}
    \operatorname{MAT}(S_g, S_r) = \textstyle\frac{1}{|S_r|} \sum_{\boldsymbol{R} \in S_r} \min_{\hat{\boldsymbol{R}} \in S_g} \operatorname{RMSD}(\boldsymbol{R}, \hat{\boldsymbol{R}}),
\end{equation}
\begin{equation}
    \operatorname{RMSD}(\boldsymbol{R}, \hat{\boldsymbol{R}}) = \min_{\Phi} \left(\textstyle \frac{1}{n} \sum_{i=1}^n \|\Phi(\boldsymbol{R}_i) - \hat{\boldsymbol{R}}_i\|^2 \right)^{\frac{1}{2}}.
\end{equation}
Here $S_g$ and $S_r$ represent the set of generated and reference conformations, respectively. We use $\hat{\boldsymbol{R}}$ to denote a generated conformation and 
$\boldsymbol{R}$ to denote a reference conformation, where
$i$ indexes heavy atoms, $n$ is the number of heavy atoms, and $\Phi$ is an optimal alignment operator.

\subsubsection{Main Results}
The results are reported in Table~\ref{tab:generation}. For brevity, we henceforth use FlexMol to denote FlexMol$^+$\textsubscript{3D}, the variant pre-trained with 3D-only data in Stage 2.
%FlexMol refers to the better variant between FlexMol$^+$\textsubscript{2D} and FlexMol$^+$\textsubscript{3D}.

\begin{table}[h]
    \centering
    \caption{Performance of molecular conformation generation on QM9. COV is measured in percent, while  MAT is measured in Å (angstroms, a length unit), representing the average atomic coordinate deviation.}
    \label{tab:generation}
    \begin{tabular}{l|cc|cc}
    \hline
     &  \multicolumn{2}{c}{COV (\%) $\uparrow$} & \multicolumn{2}{c}{MAT (Å) $\downarrow$} \\
     & Mean & Median &  Mean & Median \\
    \hline
     RDKit $[\diamond]$ & 83.26& 90.78& 0.3447& 0.2935\\
     % CVGAE &0.09 &0.00& 1.6713& 1.6088\\
     GraphDG $[\diamond]$ &73.33& 84.21& 0.4245 &0.3973\\
     CGCF $[\diamond]$ & 78.05 & 82.48 & 0.4219& 0.3900 \\
     ConfVAE $[\diamond]$ & 80.42 & 85.31 & 0.4066 & 0.3891 \\
     ConfGF $[\diamond]$ & 88.49 & 94.13 & 0.2673 & 0.2685 \\
     GeoMol $[\diamond]$ & 71.26 & 72.00 & 0.3731 & 0.3731 \\
     DGSM $[\diamond]$ & 91.49 & 95.92 & 0.2139 & 0.2137 \\
     GeoDiff $[\diamond]$& 92.65& 95.75& 0.2016 &0.2006\\
     DMCG $[\diamond]$& 94.98& 98.47& 0.2365& 0.2312\\
     \hline
     FlexMol &  \textbf{97.25} &  \textbf{100.00} & \textbf{0.1890} & \textbf{0.1741} \\
     \hline
     \rowcolor{gray!20}
     Uni-Mol $[\diamond]$ & {97.95} & {100.00}& {0.1831} &{0.1659} \\
     \hline
    \end{tabular}
\end{table}

The conformation generation results show that our model significantly outperforms existing methods such as DGSM, GeoDiff, and ConfGF. Moreover, we also achieve competitive performance with Uni-Mol, the current state-of-the-art model pre-trained on a substantially larger set of 19M molecules. This demonstrates that our model not only generates diverse molecular conformations but also maintains precision in structural alignment.

\begin{table}[h]
    \centering
    \caption{Ablation study.} 
 %\resizebox{0.9\linewidth}{!}{
    \begin{tabular}{l|cc|cc}
    \hline
    &  \multicolumn{2}{c|}{ROC-AUC $\uparrow$} & \multicolumn{2}{c}{MAE $\downarrow$}\\
    & BBBP & BACE & QM7 & QM9 \\
    \hline
    FlexMol & \textbf{75.1} & \textbf{85.7} & \textbf{52.8} & \textbf{0.00561} \\
    \hline
    w/o 3D Feature & 62.4 & 68.8 &  73.8 & 0.00764\\
    w/o 2D Feature & 70.2  & 81.1 & 54.3 & 0.00569  \\
    w/o 3D$\rightarrow$2D Decoder & 69.5 & 78.6 &  61.2 & 0.00608 \\
    w/o 2D$\rightarrow$3D Decoder & 68.6 & 75.4 & 62.3 & 0.00644\\
    w/o CS Loss & 56.7 & 70.6 & 108.1 & 0.00807\\
    w/o Rec Loss & 73.5 & 84.1 & 58.3 & 0.00559 \\
    w/o MM Encoder & 69.5 & 78.5 & 57.4 & 0.00610 \\
    \hline      
    \end{tabular}%}    
    \label{tab:ablation_study}
\end{table}

\subsection{Ablation Study}  
We further evaluate the contributions of various components in our approach. The results of the ablation study are presented in Table \ref{tab:ablation_study} for the following variants. 
\begin{itemize}[leftmargin=*]
    \item \textbf{w/o 3D Feature:} This configuration involves pre-training the model without 3D features in Stage 1 and performing continual learning using only 2D features in Stage 2. Similarly, \textbf{w/o 2D Feature} refers to pre-training without 2D features in Stage 1 and continual learning using only 3D data in Stage 2. 
    \item \textbf{w/o 2D $\rightarrow$(3D) Decoder:} In Stage 2 of the pre-training, for the missing 2D (3D) modality, the model directly utilizes the molecular representation learned by the MLP during Stage 1, without employing the decoder to generate the corresponding representation, and uses the 2D (3D)-only data for Stage 2.
    \item \textbf{w/o CS/Rec Loss:} This setup excludes the contrastive similarity (CS) loss or reconstruction (Rec) loss from the training objectives during the pre-training Stage 1 to assess their impact on the model performance.  
    \item \textbf{w/o MM Encoder:} This configuration removes the multi-modal encoder in both pre-training stages to demonstrate the effect of modality fusion on the model performance. 
\end{itemize}

The ablation results show that each component contributes to the overall performance, as removing any of them leads to a degradation in results. Removing 3D or 2D features, the contrastive similarity (CS) loss, and the MM encoder are especially critical to FlexMol’s performance,which collectively indicates that effective cross-modal alignment and fusion are central to the success of molecular representation learning.
Excluding either decoder leads to consistent drops, especially on regression tasks, highlighting the need for cross-modality translation.

However, not all components contribute equally to the performance. Specifically, the removal of the reconstruction loss (w/o Rec Loss) results in relatively smaller performance declines. This may be because in  Stage 1 of the pre-training, the output representations from the 2D$\rightarrow$3D (3D$\rightarrow$2D) decoders are used as inputs to the multi-modal encoder, and the prediction head serves as an additional loss, which already guides the molecule and atom-pair representations towards an optimal form, thus alleviating the reliance on the reconstruction loss for further refinement.

\begin{table}[tbp]
\centering
\caption{Effect of decoders in Stage 1.}
\label{tab:flexmol_ablation_decoder}
\begin{tabular}{lcc}
\hline
& ROC-AUC $\uparrow$ &MAE $\downarrow$ \\
 & BBBP  & QM9  \\
\hline
FlexMol-Stage1               & 72.3 & 0.00587 \\
FlexMol-Stage1 w/o decoder & 72.7 & 0.00574 \\
FlexMol-Stage1-3D-only       & 69.5 & 0.00639 \\
\hline
\end{tabular}

\end{table}

To further investigate the impact of the decoders in Stage 1, we observe that the usage of decoders may lead to a slight performance degradation, as their primary role is to facilitate cross-modal alignment rather than to directly optimize for downstream tasks. To assess this effect, we conduct an additional experiment on the PCQM4Mv2 dataset for 3 model variants, and use 3D-only data for all variants as the second pre-training stage. The downstream results are shown in Table \ref{tab:flexmol_ablation_decoder}. 

The results indicate that, without the decoder in Stage 1, slightly better performance than the full model can be achieved, and both models outperform the 3D-only variant by a notable margin. These findings suggest that the performance drop potentially caused by the decoders is relatively minor compared to the loss incurred from the absence of modality-specific information. More importantly, decoders are crucial in Stage 2. As Table~\ref{tab:ablation_study} shows, removing the decoder significantly harms performance (e.g., removing the 2D decoder drops ROC-AUC on BBBP from 75.1\% to 69.51\%). Thus, despite the minor Stage 1 degradation, decoders are essential for flexible modality handling and strong downstream performance.

\subsection{Effect of Single-Modality Data Size}

\begin{figure}[tbp]
\centering
\subfigure[BBBP on 3D-only data]{
    \begin{minipage}{0.43\linewidth}
    \label{fig:3d_only_data_size}
    \centering
    \includegraphics[height=2.8cm]{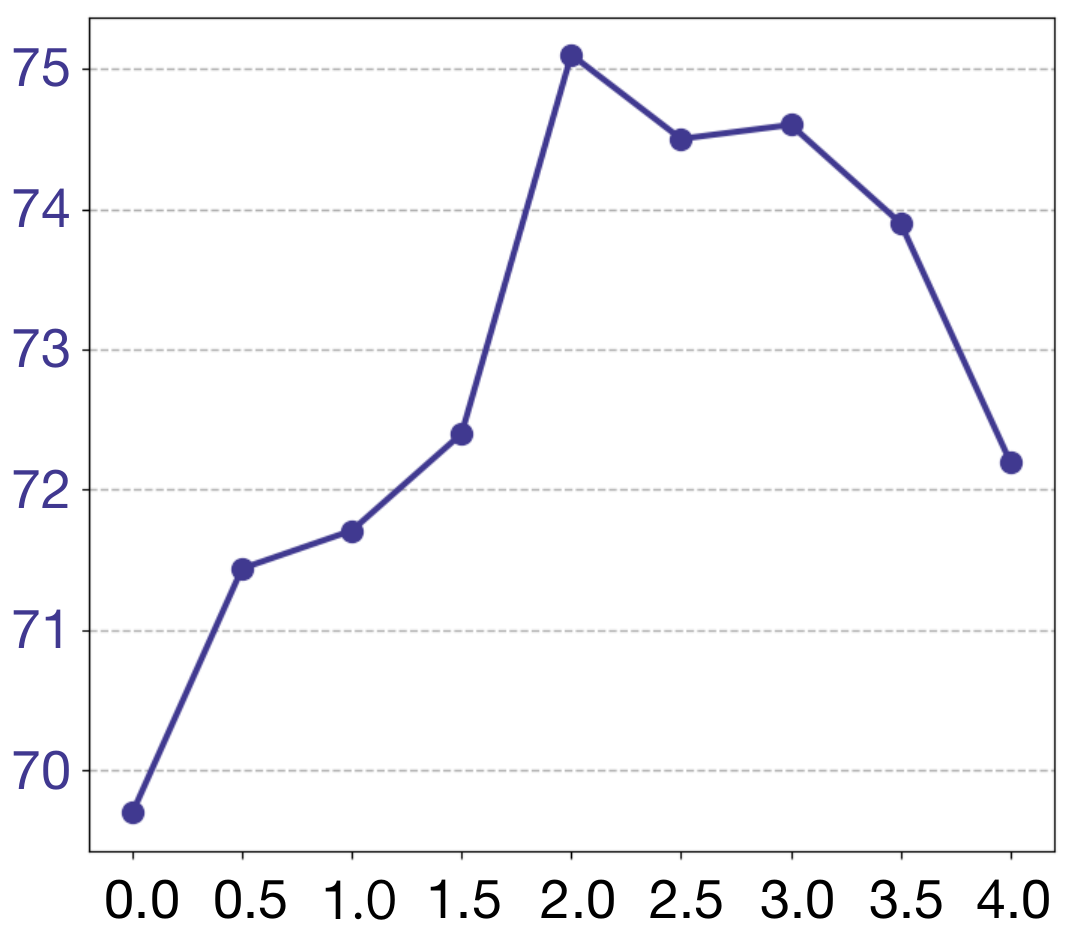}
    \end{minipage}
    }
\subfigure[BBBP on 2D-only data]{
    \begin{minipage}{0.41\linewidth}
    \label{fig:2d_only_data_size}
    \centering
    \includegraphics[height=2.8cm]{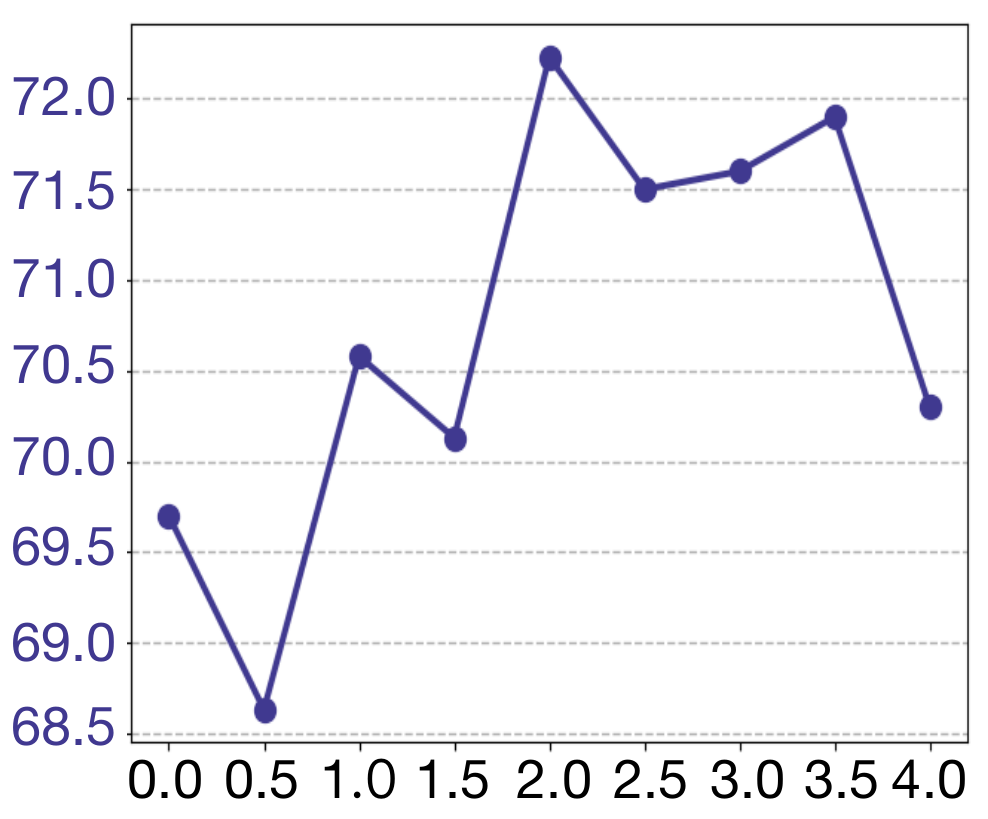}
    \end{minipage}
    }

\subfigure[QM7 on 3D-only data]{
    \begin{minipage}{0.42\linewidth}
    \label{fig:3d_only_data_size_2}
    \centering
    \includegraphics[height=2.8cm]{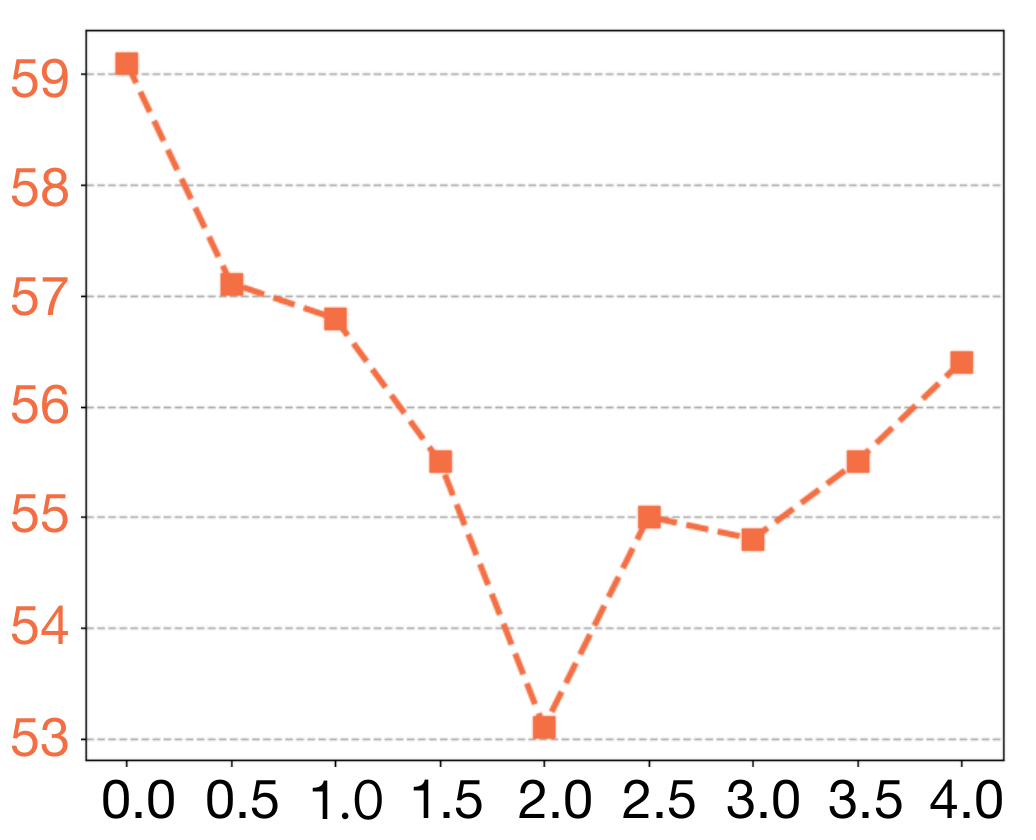}
    \end{minipage}
    }
\subfigure[QM7 on 2D-only data]{
    \begin{minipage}{0.4\linewidth}
    \label{fig:2d_only_data_size_2}
    \centering
    \includegraphics[height=2.8cm]{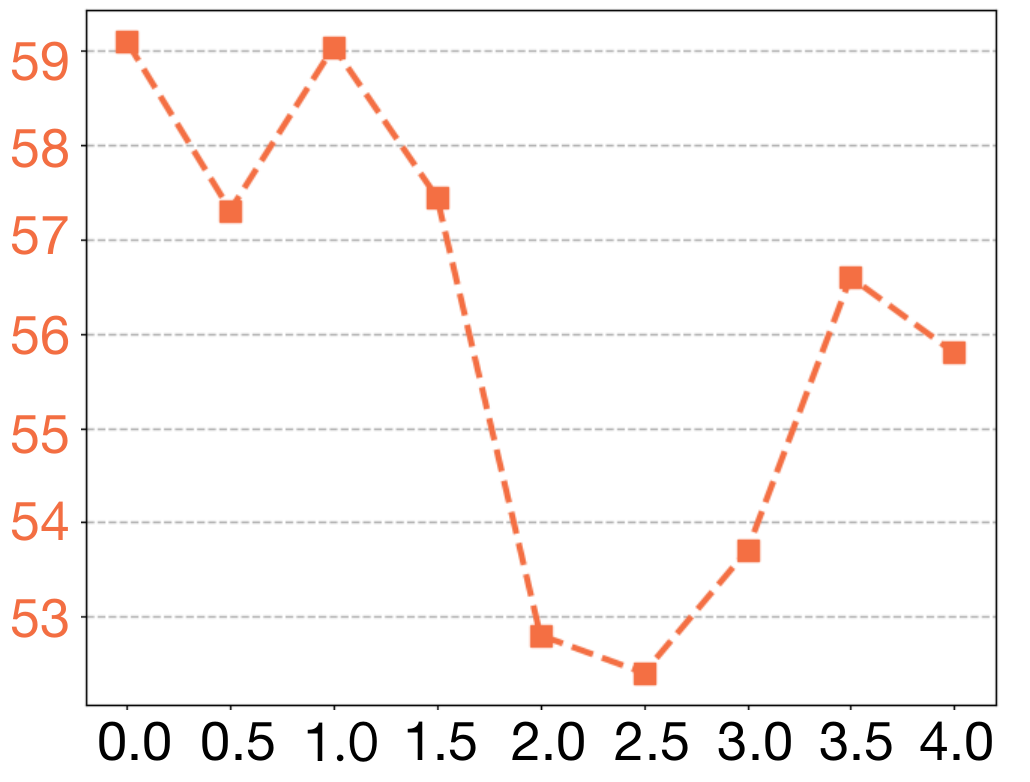}
    \end{minipage}
    }
\caption{FlexMol performance on various sizes of 2D/3D-only data in Stage 2. BBBP is evaluated in ROC-AUC ($\uparrow$) and QM7 is evaluated in MAE ($\downarrow$).
}
\label{fig:data_size_exp}
\end{figure}

% \begin{table}[H]
% \centering
% \caption{FlexMol performance on various size of 2d/3d only data.}
% \label{tab:data_size}
% \begin{tabular}{l|cc|cc}
% \toprule
% & \multicolumn{2}{c}{\textbf{3D Only}} & \multicolumn{2}{c}{\textbf{2D Only}} \\
%  & \textbf{BBBP} & \textbf{QM7} & \textbf{BBBP} & \textbf{QM7} \\
% \midrule
% 0M & 69.7 & 59.1 & 69.7 & 59.1 \\
% 0.5 M & 71.440 &
% 57.11 & 68.63 &  57.31\\
% 1M & 71.71 &
% 56.79& 70.58 & 59.04449  \\
% 1.5M & 72.4 & 55.5 & 70.13& 57.446 \\
% 2M & 75.1 & 53.1 & 72.23 &  52.8\\
% \bottomrule
% \end{tabular}
% \end{table}
We study the impact of single-modality data size on model performance during Stage 2 of the pre-training. The results for 2D-only and 3D-only data on the BBBP and QM7 datasets are shown in Figure~\ref{fig:data_size_exp}.

% When dataset sizes ranging from 0M to 2M, the model demonstrates consistent improvement as the data size increases. In the 3D-only setting, performance on both 3D and 2D molecular property predictions improves.
% A similar trend is observed in the 2D-only setting for BBBP, 
% and for QM7, MAE fluctuates but ultimately reaches the best at about 2M. This suggests that when the single-modality dataset is relatively small, the model can effectively perform continual learning on the additional single modal data.
When data sizes range from 0M to 2M, the model improves consistently with more data. In the 3D-only setting, performance on both 3D and 2D molecular property predictions improves. This suggests that the model can effectively perform continual learning initially on a moderate amount of additional single-modal data.

% These findings highlight the varying impact of modality-specific data sizes on different tasks and underscore the importance of maintaining a balanced dataset in multi-modal learning.

However, for data sizes beyond 2M, performance gains plateau. We further observe that the performance starts to decline when the  size grows toward 4M. This trend may be attributed to the extent of Stage 1 pre-training, where the upper bound of performance gains on single-modal data is constrained by the amount of paired data used in Stage 1. Once the Stage 2 data size approaches the paired data size (3.4M) in Stage 1, the model may begin to overfit to single-modal data, resulting in performance degradation.

\subsection{Hyper-parameter Sensitivity}

% unimol2d_400hop v0: 400hop on pcqm 200hop
We conduct hyper-parameter sensitivity experiments to evaluate the impact of Transformer encoder/decoder layers (denoted as `layer') and the maximum number of hops (denoted as `max\_hop') used in calculating shortest path features. The results are shown in Table~\ref{tab:hyperparameter}.  Note that since the max\_hop parameter pertains to 2D molecular features, the provided results are based on pre-training with 2D-only data in Stage 2.

\begin{table}[tbp]
    \centering
    \caption{Hyper-parameter sensitivity.} 
 
    \begin{tabular}{l|c|cc|cc}
    \hline
    & & \multicolumn{2}{c|}{ROC-AUC $\uparrow$} & \multicolumn{2}{c}{MAE $\downarrow$}\\
    & Model size & BBBP & BACE & QM7 & QM9 \\
    \hline
    layer=4 & 62.5M&72.3 &70.6 & 54.3 & 0.00569 \\
    layer=6 &87.7M &72.5 & 83.2 &55.5 & 0.00553  \\
    layer=8 & 112M &75.1 & 85.7 & 52.8 & 0.00561 \\
    \hline
    dim=512 & 46.8M& 72.0 &83.9 & 52.2 &0.00568 \\
    dim=1024 &68.9M & 71.9 & 82.5 & 53.6 & 0.00565\\
    dim=2048 & 112M & 75.1 & 85.7 & 52.8 & 0.00561 \\
    \hline
    max\_hop=100 & 112M & 72.9 & 85.7 & 53.1 & 0.00561 \\
    max\_hop=200& 114M &70.1 & 82.8 & 56.2 & 0.00569\\
    max\_hop=400 & 117M &74.0 & 83.4 & 57.0 & 0.00574 \\
    % \midrule
    % max\_hop=100 (2d) & & & \\
    % max\_hop=200 (2d)& & & \\
    % max\_hop=400 (2d) & & & \\
    \hline      
    \end{tabular}
    \label{tab:hyperparameter}
\end{table}

The analysis shows that the overall impact of hyper-parameters on performance is relatively minor. Reducing layers or dimension size does not significantly degrade results, indicating that the model is somewhat robust to parameter reduction. This suggests the potential for lighter-weight models that balance efficiency with minimal performance loss in the future. Additionally, increasing the hop count does not yield performance improvements, which suggests that beyond a certain point, further increasing the hop count may not contribute to better performance and could even introduce unnecessary complexity.

\subsection{Computational Efficiency}

We compare the computational efficiency of FlexMol, its variants without modality decoders, and Uni-Mol, all trained on the 3.4M-scale PCQM4Mv2 dataset. The results are shown in Table~\ref{tab:combined_comparison_transposed}. Here, \#Params denotes the number of model parameters, $T_{\text{train}}$/eph represents the average training time per epoch (only Stage 1 for FlexMol and its variants), $T_{\text{infer}}$ is the average inference latency per sample, and GPU Mem indicates the peak memory usage during training. For multi-GPU settings, the memory usage per GPU is reported alongside the number of GPUs used (e.g., 24 $\times$ 2). All measurements are conducted under the same dataset and batch size settings to ensure fairness.

\begin{table}[ht]
\centering
%\addtolength{\tabcolsep}{-1mm}
\caption{Computational efficiency comparison of FlexMol, its variants w/o modality decoders, and Uni-Mol.}
\label{tab:combined_comparison_transposed}
\resizebox{\linewidth}{!}{
\begin{tabular}{lcccc}
\toprule
\textbf{Metric} & \textbf{FlexMol} & \textbf{w/o 2D$\rightarrow$3D} & \textbf{w/o 3D$\rightarrow$2D} & \textbf{Uni-Mol} \\
\midrule
\#Params (Millions) & 112  & 47  & 47  & 45.5  \\
$T_{\text{train}}$/eph (GPU hrs) & 7  & 4.5  & 4  & 4  \\
$T_{\text{infer}}$ (ms) & 28  & 25  & 20  & 18  \\
GPU Mem (GB) & 24$\times$2 & 22$\times$2 &  39 &  36 \\
\bottomrule
\end{tabular}
}
\end{table}

The results show that the introduction of modality decoders results in moderate computational overhead in terms of training time, inference latency, and parameter count. However, this overhead is justified by the observed performance improvement, making it a reasonable trade-off for practical deployment.
Futhermore, with the adoption of a parameter-sharing mechanism in the self-attention layers, the trainable parameter count is reduced from 248M to 112M, significantly improving memory and computational efficiency.

\section{Conclusion and Future Work}
In this work, we propose a unified framework for molecular pre-training that addresses the limitations of existing methods in leveraging both 2D and 3D molecular data. Our approach effectively integrates single and paired modality inputs, enabling flexible learning scenarios. By combining separate models for 2D and 3D data with shared parameters, we achieve the fusion of modality-specific representations while maintaining computational efficiency. 
The proposed decoders further enhance the capability of the framework by generating missing modality data, ensuring robust multi-modal learning even with single-modality inputs. Extensive experiments show that our approach delivers strong performance on diverse molecular property prediction and conformation generation tasks, surpassing existing models trained on smaller or similar-scale datasets, while remaining competitive with large-scale state-of-the-art pre-trained models.

Our model still has certain limitation that merit further investigation. The scalability of the model is constrained by the limited availability of paired data. In the second pre-training stage, where single-modal data is introduced, performance improvements remain dependent on the paired data from the first stage. When the amount of single-modal data substantially exceeds that of paired data, performance deteriorates due to potential overfitting to single-modal data,
which restricts scaling to larger single-modal datasets.
Future work will therefore focus on 
balancing paired and single-modal data, as well as constructing higher-quality paired datasets to enhance modality alignment and enable large-scale learning.

\section*{Acknowledgments}
This research / project is supported by the Ministry of Education, Singapore under its Academic Research Fund (AcRF) Tier 1 grant (22-SIS-SMU-054). Any opinions, findings and conclusions or recommendations expressed in this material are those of the author(s) and do not reflect the views of the Ministry of Education, Singapore.

\section*{GenAI Usage Disclosure}
Portions of this manuscript were refined with the assistance of OpenAI's ChatGPT. All AI-generated content was reviewed and revised by the authors. The final responsibility for the content rests solely with the authors.

\bibliographystyle{ACM-Reference-Format}
\balance
\bibliography{ref}

\end{document}